\documentclass[twoside,11pt]{article}
\usepackage[preprint]{jmlr2e}

\usepackage{lastpage}

\ShortHeadings{Bayesian Pseudo Posterior Mechanism for Differentially Private ML}{Chew, Williams, Segarra, Preiss, Konet, Savitsky}
\firstpageno{1}

\usepackage{comment}
\usepackage{bm}
\usepackage{graphicx,psfrag,epsf}
\usepackage{enumerate}
\usepackage{natbib}
\usepackage{comment}
\usepackage{booktabs}
\usepackage{longtable}
\usepackage{multirow}
\usepackage{setspace}
\usepackage{commath}
\usepackage{url} 

\begin{document}


\title{\bf Bayesian Pseudo Posterior Mechanism for Differentially Private Machine Learning}

\author{\name Robert Chew \email rchew@rti.org \\
       \addr Center for Data Science and AI\\
       RTI International \\
       Research Triangle Park, NC 27709, USA
       \AND
       \name Matthew R. Williams \email mrwilliams@rti.org \\
       \addr Center for Official Statistics\\
       RTI International\\
       Washington, DC 20005, USA
       \AND
       \name Elan A. Segarra \email segarra.elan@bls.gov \\
       \addr Office of Compensation and Working Conditions\\
       U.S. Bureau of Labor Statistics\\
       Hillcrest Heights, MD 20746, USA
       \AND
       \name Alexander J. Preiss \email apreiss@rti.org \\
       \addr Center for Data Science and AI\\
       RTI International \\
       Research Triangle Park, NC 27709, USA
       \AND
       \name Amanda Konet \email akonet@rti.org \\
       \addr Center for Data Science and AI\\
       RTI International \\
       Research Triangle Park, NC 27709, USA
       \AND
       \name Terrance D. Savitsky \email savitsky.terrance@bls.gov \\
       \addr Office of Survey Methods Research\\
       U.S. Bureau of Labor Statistics\\
       Hillcrest Heights, MD 20746, USA
       }

\maketitle

\begin{abstract}
Differential privacy (DP) is becoming increasingly important for deployed machine learning applications because it provides strong guarantees for protecting the privacy of individuals whose data is used to train models.  However, DP mechanisms commonly used in machine learning tend to struggle on many real world distributions, including highly imbalanced or small labeled training sets. In this work, we propose a new scalable DP mechanism for deep learning models, \texttt{SWAG-PPM},  by using a pseudo posterior distribution that downweights by-record likelihood contributions proportionally to their disclosure risks as the randomized mechanism. As a motivating example from official statistics, we demonstrate SWAG-PPM on a workplace injury text classification task using a highly imbalanced public dataset published by the U.S. Occupational Safety and Health Administration (OSHA). We find that SWAG-PPM exhibits only modest utility degradation against a non-private comparator while greatly outperforming the industry standard DP-SGD for a similar privacy budget. 
\end{abstract}


\section{Introduction}
\label{sec:intro}

Machine learning has become ubiquitous in modern society, with deployed models capable of processing vast amounts of personal and potentially sensitive data. This widespread adoption has raised significant privacy concerns, particularly regarding the protection of individual information used during model training. Differential privacy (DP) \citep{dwork2006differential} has emerged as a crucial framework for addressing these concerns by providing mathematical guarantees about the privacy of training data while maintaining model utility.

Despite its theoretical appeal and growing adoption in industry, implementing DP in practical machine learning applications presents several challenges. Current DP mechanisms, while effective in idealized settings, often suffer under real-world complications such as highly imbalanced datasets \citep{rosenblatt2024differential} or limited labeled training data \citep{suriyakumar2021chasing} that are common to domains like healthcare, finance, and government.

The most widely adopted approach for training deep learning models with differential privacy is DP-SGD (Differentially Private Stochastic Gradient Descent) \citep{10.1145/2976749.2978318}, which adds calibrated noise to gradients during training. However, DP-SGD faces several significant limitations, particularly when applied to imbalanced datasets. \citet{bagdasaryan2019differential} demonstrate empirically that when DP-SGD is applied to highly imbalanced data, the less represented groups, which already exhibit lower accuracy, experience a greater loss in utility. Furthermore, they find this gap widens as stricter privacy guarantees are enforced. \citet{farrand2020neither} extends these investigations of DP-SGD under a wider range of subgroup imbalance and privacy budget.  They find that degradation of model accuracy is not only limited to highly imbalanced data, but can also occur in situations where classes are slightly imbalanced. They also find that these disparate impacts can be observed even under loose privacy guarantees. In an applied healthcare study, \citet{suriyakumar2021chasing} found that DP-SGD tends to lose critical information about minority groups, such as dying patients or ethnic minorities, effectively ignoring those in the tails of the data distribution. Moreover, when DP-SGD has been proposed to extend prior imbalanced learning methods with a differential privacy guaruntee (``private imbalanced learning''), its has also faced challenges.  For example, \citet{rosenblatt2024differential} tests the well established method of class weights (inversely proportional to class size) to improve imbalanced learning, once training with SGD as a non-private benchmark and again with DP-SGD for privacy protection. They find that while the non-private class-weighted models exhibited strong predictive performance as expected, the private class-weighted version using DP-SGD severely underperformed. They conclude that methods designed to address class imbalances using DP-SGD for privacy guarantees may not be ideal for small to medium-sized datasets.

In this paper, we introduce SWAG-PPM, a novel differentially private mechanism that takes a fundamentally different approach. Rather than relying on additive noise, our method achieves privatization through the use of an approximate posterior distribution sampled across SGD epochs as the randomized mechanism. This approach induces distortion through parameter smoothing rather than explicit noise addition, offering more nuanced privacy protection that better preserves model utility, particularly for imbalanced datasets.

\subsection{Motivation}
Highly parameterized deep neural network models are increasingly used by government statistical agencies to classify text data obtained from survey respondents into preset categories used to quantify and publish associated statistics.  Our paper is motivated by the ``autocoder'' model developed at the U.S. Bureau of Labor Statistics (BLS) that classifies characteristics of illness and injury cases provided by the Survey of Occupational Injuries and Illnesses (SOII) administered to business establishments \citep{auto}.  Each responding establishment submits a ``narrative'' for each injury or illness event that resulted in one or more days of missed work, job reassignment, or restrictions in job responsibilities.  A narrative would include the affected employee's job title, a description of what the employee was doing just before the incident, a description of the incident or illness, and what object or substance may have directly harmed the employee.  These narrative variables are coded by the autocoder into categories defined by BLS (e.g., coding a job title into a Standard Occupational Classification code). 

The BLS autocoder uses a distilled version of the RoBERTa transformer architecture as its pre-trained model, first introduced in \citet{liu2019robertarobustlyoptimizedbert}. The estimation algorithm utilizes a variant of stochastic gradient descent, where a randomly-selected small-sized (e.g., $8-16$ records) ``mini-batch'' or collection of records is used to approximate the full data average gradient on each iteration step. Gradient updates are often performed over multiple passes through the entire training dataset, referred to as ``epochs''.

Along with most other U.S. statistical agencies, BLS is subject to the Confidential Information Protection and Statistical Efficiency Act (CIPSEA) which requires BLS to protect the identities and data submitted by survey participants in the SOII.  Yet, partner agencies and other interested parties often desire access to the trained BLS autocoder to perform classifications on their own data.  Our goal is to privatize the estimated parameters of the autocoder model trained on the closely-held SOII data to facilitate its sharing with partner agencies as well as external researchers and policy-makers. 

\subsection{Literature Review}
 In this work, we focus on the differential privacy (DP) guarantee that is generally defined as:
\begin{definition}
    A randomized mechanism $\mathcal{M}: \mathcal{D}^{n} \rightarrow \mathcal{R}$ with domain $\mathcal{D}^{n}$ defined as the space of datasets of size $n$ and range $\mathcal{R}$ satisfies $(\epsilon,\delta)-$ differential privacy (DP) if for any two datasets, $D$ and $D'$ that differ by a single record and for any output set $S \in \mathcal{R}$ we have the following result:
    \begin{equation}
        \Pr\left( \mathcal{M}(D) \in S\right) \leq \epsilon \times \Pr\left( \mathcal{M}(D') \in S\right) + \delta,
    \end{equation}
\end{definition}
where $\delta\in [0,1]$ is interpreted as the probability of achieving a ``bad'' outcome that produces an $\epsilon^{\ast} > \epsilon$.  Any mechanism with $\delta > 0$ is defined as ``relaxed'' in the sense that a bad outcome is not accounted for in the privacy guarantee.  The case of $\delta$ is equal to $0$ almost surely under the mechanism is defined as ``strict'' differential privacy.

For example, let $f(D)$ denote some real function of dataset, $D$, and define the sensitivity of $f$, $\Delta$, to be the supremum $\abs{f(D)-f(D')}$ over all neighboring datasets $D, D'\in \mathcal{D}$, where neighboring datasets differ by one record.  The Gaussian mechanism is defined as $\mathcal{M}(D) = f(D) + \mathcal{N}\left(0,(\Delta\sigma)^2\right)$ and achieves $(\epsilon,\delta)-$DP
if the privacy parameter $\delta$ satisfies the condition given in \citet{10.1145/2976749.2978318}, such as $\delta \geq \frac{4}{5} \exp(-(\sigma \epsilon)^2 / 2)$. The degree of noise introduced (distortion) is determined by the sensitivity of $f$ and the targeted privacy parameters $(\epsilon, \delta)$. Stricter privacy guarantees, characterized by smaller values of $\epsilon$ and/or $\delta$, require more noise to be added to ensure privacy. This noise is proportional to the sensitivity of the function, which measures the maximum possible change in $f$ due to the addition or removal of a single record \citep{dwork2006differential}.  

Constructing a differentially private mechanism for models trained using an SGD estimation algorithm with an additive noise mechanism involves applying the mechanism at the mini-batch level. The mechanism is analyzed as the composition of a sequence of additive noise mechanisms, each corresponding to a mini-batch update in the SGD algorithm \citep{10.1145/2976749.2978318}. The privacy guarantee, $\epsilon$, for the entire SGD algorithm is bounded by the composition of the privacy guarantees of the mini-batch mechanisms. Since each mini-batch contributes to privacy exposure, the overall privacy guarantee for the SGD algorithm depends on the number of mini-batch updates and the specific composition method used.

\citet{song} construct a mechanism for each iteration (or training step) of the SGD algorithm by adding Gaussian noise to the gradients computed on mini-batches. The noise scale is proportional to  step-size and the gradient norm and inversely proportional to the batch size and the privacy parameter $\epsilon_{t}$, where $t$ indexes the mini-batch. The overall $(\epsilon, \delta)$-DP guarantee for the SGD algorithm is derived using naïve composition: $\epsilon = \sum_{t=1}^{T} \epsilon_t$, where $T$ represents the total number of iterations required for the parameters to converge to a solution. This additive composition provides a relatively loose bound, requiring a higher noise scale compared to more sophisticated sub-additive methods such as advanced composition or moments accountant techniques. These tighter methods reduce the growth of $\epsilon$, potentially leading to improved privacy guarantees as the number of iterations $T$ increases.

\citet{10.1145/2976749.2978318} construct a tighter upper bound for the privatized SGD mechanism by analyzing the summation of upper bounds defined on moment generating functions parameterized by $\alpha$, where each bound corresponds to a training iteration.  The bounded moments is converted to $(\epsilon,\delta)-$DP by using a tail result that constructs $\delta$ by minimizing the moment bound achieved from the summation of the per-iteration bounds over the moment parameter, $\alpha$. This approach achieves a significantly tighter privacy bound compared to naïve composition. Specifically, their method achieves a sub-additive bound for the SGD mechanism, with an $\epsilon$ value that is reduced by a factor of approximately $\log(T / \delta)$ compared to the direct composition of per-iteration guarantees. This demonstrates the effectiveness of optimizing privacy loss via moments accountant techniques in improving privacy guarantees for differentially private SGD.

\citet{Mironov2017RnyiDP} generalize the moments accounting approach of \citet{10.1145/2976749.2978318} by bounding the $\alpha$-Rényi divergence $D_{\alpha}(\mathcal{M}(D) \Vert \mathcal{M}(D'))$ of a mechanism $\mathcal{M}$, ensuring that $D_{\alpha}(\mathcal{M}(D) \Vert \mathcal{M}(D')) \leq \epsilon$. This formulation introduces Rényi differential privacy (RDP), a privacy guarantee parameterized by $(\alpha, \epsilon)$. As $\alpha \to \infty$, the Rényi divergence converges to the supremum divergence, aligning with strict $\epsilon$-DP (where $\delta = 0$). The $(\alpha, \epsilon)$-RDP framework provides a more flexible, relaxed form of differential privacy that can be converted into the usual $(\epsilon, \delta)$-DP using tailored bounds based on the chosen $\delta$.

\citet{Mironov2017RnyiDP} also demonstrate that the Gaussian mechanism with standard deviation $\sigma$ satisfies $(\alpha, \alpha / \sigma^2)$-RDP, where $\epsilon$ grows linearly with $\alpha$. They further show that $\alpha$ can be optimally chosen to minimize the overall privacy loss when composing multiple RDP mechanisms, such as those arising from iterative updates in stochastic gradient descent (SGD). This optimization achieves a sub-additive privacy bound, providing tighter guarantees compared to naïve composition.

\citet{mironov2019renyi} define their ``sampled'' Gaussian mechanism by embedding a randomized selection step into a Gaussian mechanism that operates on mini-batches within each epoch of SGD.  They evaluate the statistic $f$ (such as gradient updates) on a randomly selected subset of the dataset, where the selection step is typically uniform, and then add Gaussian noise to ensure differential privacy. This composite $(\alpha,\epsilon)-$ RDP mechanism, along with the $\alpha$-optimized composition of mechanisms over the SGD training steps, as implemented in \citet{li2021large}, is used in our DP-SGD benchmark comparison. 

In contrast, \citet{dimitrakakis2017differential} introduce a Bayesian mechanism for achieving differential privacy by leveraging the posterior distribution of a hierarchical probability model, $\mathcal{M}(D,\theta) = \xi(\theta\mid D)$.  The mechanism defines a sensitivity statistic as the supremum over $\theta \in\Theta$ and neighboring databases $D, D'\in\mathcal{D}$ of $\abs{\ell_{\theta}(D) - \ell_{\theta}(D')}$ where $\ell_{\theta}(\cdot)$ denotes the model log-likelihood. \citet{JMLR:v23:21-0936} extends this approach with their pseudo posterior mechanism to make it practical by using record indexed weights, $\alpha_{i} \in [0,1]$, that downweight each record $D_{i}$ in proportion to its empirical privacy risk defined as $1/\abs{\ell_{\theta}(D_{i})}$. By reducing the influence of high-risk records, this approach aims to control sensitivity and ensure it remains finite.

To extend the use of the pseudo posterior mechanism to a deep learning regime, we adopt the Stochastic Weight Averaging - Gaussian (SWAG) approach from \citet{maddox2019simple}. SWAG treats parameters estimated across epochs as though they are draws from an approximate multivariate Gaussian posterior distribution, relying on the Bayesian central limit theorem for the convergence of the joint posterior to a Gaussian. This approach is compatible with industry standard SGD implementations and can operate in high-dimensional parameter spaces, where direct posterior estimation can be computationally prohibitive. Its flexibility and extensibility also allow for adaptations to accommodate specific modeling requirements, including privacy-preserving constraints or other domain-specific adjustments.

\section{SWAG Pseudo Posterior Mechanism}\label{sec:methods}

At its core, our method extends an existing differential privacy algorithm to integrate with modern deep learning models. The Pseudo Posterior Mechanism (PPM) (Section \ref{methods-ppm}) provides a formally private mechanism by sampling from a perturbed posterior distribution, ensuring differential privacy guarantees. SWAG (Section \ref{methods-swag}), on the other hand, offers a method to estimate the posterior distribution of a trained machine learning model efficiently, without requiring full Bayesian sampling. Together, these components enable us to construct a mechanism, the SWAG Pseudo Posterior Mechanism (SWAG-PPM) (Section \ref{methods-swag-ppm}) that combines formal privacy guarantees with scalable and practical posterior approximation for deep learning models.

\subsection{Pseudo Posterior Mechanism (PPM)}
\label{methods-ppm}
Let our deep learning parameters be represented by the parameter vector $\theta$.
We recall that the posterior distribution for $\theta$ is proportional to the product of the model likelihood and the prior measure:
\begin{equation}
\label{postmech}
\xi(\theta \mid D) \propto \mathop{\prod}_{i=1}^{n}p(D_{i}\mid \theta) \times \xi(\theta),
\end{equation}
\citet{dimitrakakis2017differential} shows that a formal privacy bound for a single draw $\theta$ from the posterior distribution \eqref{postmech} can be obtained from the log-likelihood $\ell_{\theta}(D_i) = \log p(D_{i}\mid \theta)$ and its sum $\ell_{\theta}(D) = \sum_{i}^{n} \log p(D_{i}\mid \theta)$. Then the sensitivity bound is \\$\Delta = \mathop{\sup}_{D\in \mathcal{D}^{n}:\delta(D, D') = 1}  \mathop{\sup}_{\theta \in \Theta} | \ell_{\theta}(D) - \ell_{\theta}(D') |$,
which is a supremum over the support $\Theta$ and over all databases $D$ of size $n$ and all neighboring databases $D'$. Each draw from $\theta \sim \xi(\theta \mid D)$ has a privacy bound of $\epsilon = 2 \Delta$.
We define a neighbor based on the leave-one-out (LOO) definition. Then $D' \in \mathcal{D}^{n-1}$ such that $D'$ is a proper subset of $D$ and only one record in $D$ is missing in $D'$, calling this $\delta(D, D') = 1$. Under this LOO neighborhood and the conditional independence of the likelihoods in \eqref{postmech}, \cite{JMLR:v23:21-0936} simplify the sensitivity
$\Delta = \mathop{\sup}_{d\in \mathcal{D}^{1}} \mathop{\sup}_{\theta \in \Theta} | \ell_{\theta}(d)| =  \mathop{\sup}_{d\in \mathcal{D}^{1}} \Delta_d$,
which maximizes over the individual record-level risk $\Delta_d$.

When the sensitivity $\Delta$ is finite, \citet{wang2015privacy} note and \citet{JMLR:v23:21-0936} further develop the connection to the exponential mechanism of \citet{mcsherry2007mechanism}. In order to obtain an arbitrary $\epsilon \ne 2 \Delta$, the simplest and most common approach is to scale the utility function (in our case the log-likelihood) by a factor $\alpha = \epsilon/2\Delta$. Unfortunately, when the original unweighted $\Delta = \infty$, as is the case for many common statistical models (e.g. Gaussian), this scalar weighting approach will not work. Instead, \citet{JMLR:v23:21-0936} propose the following vector-weighted pseudo-posterior mechanism (PPM):
\begin{equation}
\label{pseudomech}
\xi^{\bm{\alpha}(D)}(\theta \mid D) \propto \mathop{\prod}_{i=1}^{n}p(D\mid \theta)^{\alpha_{i}} \times \xi(\theta),
\end{equation}
where the $\alpha_{i} \in [0,1]$ serve to downweight the likelihood contributions such that highly risky records are more strongly downweighted. We must choose some decreasing function $m(): r_i \mapsto [0,1]$ which maps the risk measure $r_i = \mathop{\sup}_{\theta \in \Theta} \abs{\ell_{\theta}(D_{i})} = \Delta_i$ to a set of weights in $[0,1]$.  We choose that $m(0) = 1$ and $m(\infty) = 0$. This allows for many records to receive minimal to no downweighting, while others will be maximally downweighted towards 0 for large risk $\Delta_i$. This differential weighting scheme ensures a finite sensitivity defined as:
$\Delta_{\bm{\alpha}}  = \mathop{\sup}_{D\in \mathcal{D}^{n}:\delta(D, D') = 1}  \mathop{\sup}_{\theta \in \Theta} | \ell^{\alpha}_{\theta}(D) - \ell^{\alpha}_{\theta}(D') | < \infty$.  Similarly, we can simplify the sensitivity as maximizing over individual risks: 
$\Delta_{\alpha} = \mathop{\sup}_{d\in \mathcal{D}^{1}} \mathop{\sup}_{\theta \in \Theta} | \alpha(d) \ell_{\theta}(d)| =  \mathop{\sup}_{d\in \mathcal{D}^{1}} \Delta^{\alpha}_d$

In practice, the PPM approach of \cite{JMLR:v23:21-0936} evaluates the sensitivity $\Delta_{\alpha,D}$ on a specific database $D$ and its neighbors $D'$. This leads to a so-called local privacy bound $\epsilon_D = 2 \Delta_{\alpha,D}$. They show that this local $\Delta_{\alpha,D}$ contracts on the global $\Delta_{\alpha}$ asymptotically as the size $n$ of the database grows. This local DP approach combined with an asymptotic global guarantee is a relaxed form of DP, which they call `asymptotic' DP (aDP). 
Conceptually this is similar to an $\epsilon,\delta$-DP result in which the $\delta \rightarrow 0$ as $n \rightarrow \infty$. See Appendix \ref{app:aDP} for more details.
\citet{hu2022mechanisms} provide an alternative for ensuring global privacy bounds ($\delta = 0$) for the PPM under finite sample size, which comes at the cost of a reduction in utility for this more stringent guarantee over all possible data sets in $\mathcal{D}^{n}$.

\subsection{Stochastic Weight Averaging - Gaussian (SWAG)}
\label{methods-swag}

Stochastic Weight Averaging - Gaussian (SWAG), introduced in \citet{maddox2019simple}, is a scalable approximate Bayesian inference method designed for deep learning models.  It builds on previous work by \citet{mandt2017} showing that SGD with a constant learning rate simulates a Markov chain with a stationary distribution, implying that it is possible to build approximate Bayesian posterior inference algorithms by repurposing model artifacts that are generated during training. SWAG works by running SGD with a constant learning rate schedule for additional epochs after an initial training regime (such as fine-tuning a pre-trained model to a new task).  These additional SGD training steps explore the loss landscape around the initial training run and capture intermediate model parameters at the end of each training epoch. The set of collected model parameters are then used to estimate an approximate joint posterior using a multivariate Gaussian distribution.

To estimate the mean of the multivariate Gaussian, SWAG simply averages the model parameters obtained after each epoch $t$ run with SGD under the constant learning rate:

\begin{equation}
\label{swag-mean}
\bar{\theta} = \frac{1}{T} \sum_{t=1}^{T} \theta_t
\end{equation}

To estimate the covariance of the multivariate Gaussian, SWAG uses an average of the diagonal covariance approximation and a low-rank approximation of the sample covariance matrix to balance computational efficiency and flexibility: $\mathcal{N}(\bar{\theta}, \frac{1}{2} \cdot (\Sigma_{\text{diag}} + \Sigma_{\text{low-rank}}))$.  

The diagonal covariance approximation is computed by:

\begin{equation}
\label{swag-cov-diag}
\Sigma_{\text{diag}} = \text{diag}(\overline{\theta^2} - \bar{\theta}^2), \quad  \text{where } \overline{\theta^2} = \frac{1}{T} \sum_{t=1}^{T} \theta_t^2
\end{equation}

The low rank approximation is estimated by:

\begin{equation}
\label{swag-cov-low-rank}
\Sigma_{\text{low-rank}} = \frac{1}{K-1} \cdot \hat{D}\hat{D}^\intercal
\end{equation}

where $D$ is the deviation matrix comprised of columns $D_t = (\theta_t - \bar{\theta_t})$, $ \bar{\theta_t}$ is the running estimate of parameters' mean obtained from the first $t$ epochs, and $\hat{D}$ is an approximation of the deviation matrix using the last $K$ columns equal to $D_t$ for $t = T - K + 1,...,T$. 

With the Gaussian distribution estimated, sampling from it corresponds to draws from an approximate posterior distribution over the model parameters. 

\subsection{SWAG Pseudo Posterior Mechanism (SWAG-PPM)}
\label{methods-swag-ppm}

\begin{figure}[h]
\centering
\includegraphics[width=\textwidth]{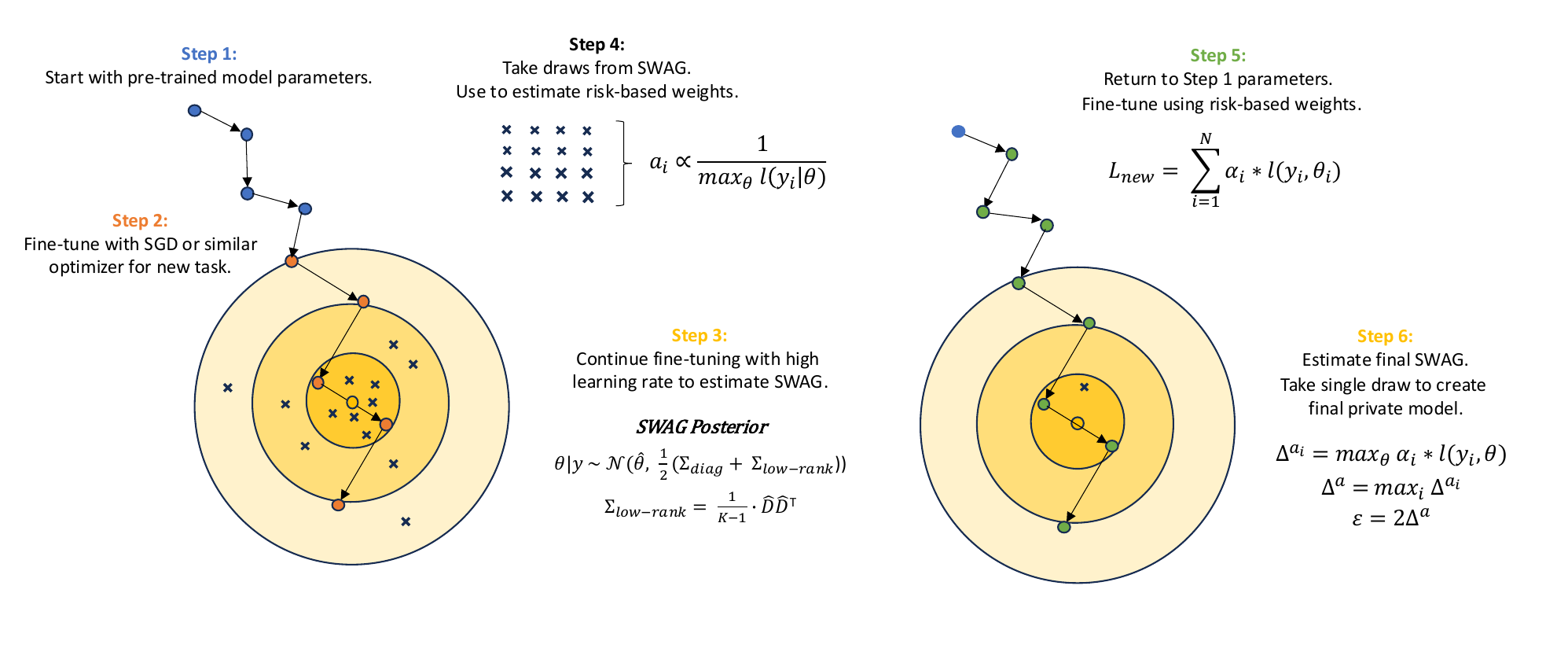}
\caption{SWAG Pseudo Posterior Mechanism}
\label{fig:swag-ppm}
\end{figure}

The SWAG Pseudo Posterior Mechanism (SWAG-PPM) combines elements of SWAG for approximate Bayesian inference and PPM for differential privacy guarantees (Figure \ref{fig:swag-ppm}). SWAG-PPM starts with fine-tuning a pre-trained model with SGD (or similar optimizer) for a new task. In our case, we use a distilled version of RoBERTa, introduced in \citet{liu2019roberta}, as our pretrained model and update the parameters to perform text classification for injury coding on new training data.  The next step in SWAG-PPM is to continue training with SGD using a constant high learning rate to estimate an initial SWAG model.  Draws from this initial SWAG model are used to calculate risk-based weights for each record as in PPM. The training process then repeats from the beginning, but instead of fine-tuning the base pre-trained model using the original likelihood function, the pseudo-likelihood function is used by incorporating the risk-based weights to downweight more risky observations.  Lastly, to create a final approximate posterior distribution, a final SWAG model is estimated using the pseudo-likelihood and risk-based weights.  A single draw from this final SWAG is used to release the private classifier with an $\epsilon_D = 2 \Delta_{\alpha,D}$.

The choice of the weight mapping $\alpha = m(r)$ may not be fully efficient if many final individual weighted risks $\Delta_{\alpha,D_i}$ are well below the maximum $\Delta_{\alpha,D}$. This means that some records may be overly penalized relative to their risk. To improve utility, we follow the iterative approach of \cite{savitsky2020re}. We begin with initial weights from the linear transformation
$\tilde{r}_{i} = \frac{r_i - \min_j r_j}{\max_j r_j - \min_j r_j}$.
We then set weights $\alpha_i =  c \times (1-\tilde{r}_{i}) + g$, where $c$ and $g$ denote scaling and shift parameters for additional tuning. We then implement a re-weighting strategy:
\begin{equation}
\alpha_i^{w} = k \times \alpha_i \times \frac{\Delta_{\bm{\alpha},D}}{\Delta_{\bm{\alpha},D_i}},
\end{equation}
where a constant, $k < 1$, is used to ensure that the final maximum Lipschitz bound remains equivalent before and after this re-weighting step. Both $\Delta_{\bm{\alpha},D}$ and $\Delta_{\bm{\alpha},D_i}$ are computed from the $\bm{\alpha}-$weighted SWAG-PPM. Thus the re-weighted mechanism SWAG-PPM requires a third round of SWAG with updated weights $\alpha_i^{w}$. \cite{savitsky2020re} demonstrate through simulation that this results in more efficient estimation (more weights being closer to 1) while still preserving the risk upper bound $\Delta_{\alpha,D}$.

Both the SWAG procedure for approximating the posterior distribution and the PPM procedure for inducing formally private draws from the posterior have asymptotic properties that are compatible with one another. For large sample sizes, the SWAG procedure's use of a multivariate Gaussian distribution for the posterior becomes more and more accurate. Similarly, the local DP privacy property becomes stronger (e.g., global) as the sample size increases. As the sample size grows to infinity and the number of parameters is constant (but still large), the risk-weighted pseudo-posterior converges to a multivariate Gaussian distribution centered at the set of parameters $\theta^*$ that minimize the Kullback-Leibler divergence to the true set of parameters $\theta_0$. Because of the risk-based down-weigthing (a type of model misspecification) this divergence is not zero. However, a single point of convergence still exists. See \cite{JMLR:v23:21-0936} for more details.

\section{Application}\label{sec:simulation}

\subsection{Data}

To demonstrate our approach, we use public data from the U.S. Occupational Safety and Health Administration (OSHA) Severe Injury Reports from January 2015 to September 2023. This dataset of severe work-related injuries reported under OSHA regulation 29 CFR 1904.39. OSHA uses version 2.01 of the Occupational Injury and Illness Classification System (OIICS) to code for ``Nature of Injury'', ``Part of Body Affected'', ``Event or Exposure'', and ``Source and Secondary Source of Injury and Illness''. The ``Final Narrative'' text field is used as model inputs and the ``Nature of Injury'' field is used as our outcome variable.  The original dataset has 86,210 observations spanning 199 unique codes with a highly imbalanced class distribution, which is common for occupational injury datasets such as SOII \citep{bls_r31}. 

\begin{figure}[h]
\centering
\includegraphics[width=\textwidth]{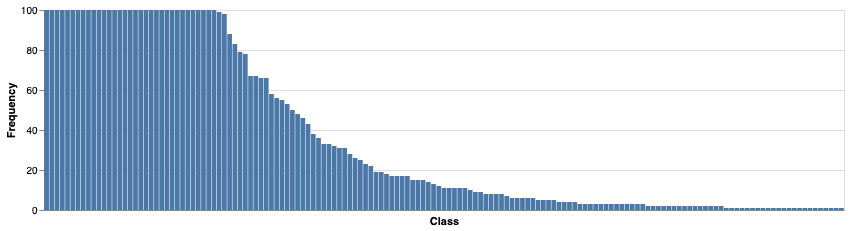} 
\caption{Training Set Class Distribution}
\label{fig:train-dist}
\end{figure}

To reduce the computational burden and cost associated with extended GPU usage while ensuring that each class in the original dataset is retained, we draw a stratified random sample by class with proportional allocation up to a maximum of 200 records. We then remove all classes with only a single observation, to ensure that each class has an opportunity to be a part of both the training and test sets. This results in 
a final analysis dataset of 10,692 observations and 153 unique codes, which we then partition into training and test sets using a stratified 50/50 split. Figure \ref{fig:train-dist} shows the resulting highly imbalanced class distribution in the training set (gini coefficient = 0.6).
The full set of codes included in the final analysis dataset, with class sizes, is available in Table \ref{tab:detailed-utility} of the Appendix.

\subsection{Model Benchmarks}

\textbf{Non-private.} The ``non-private'' benchmark model, while providing no privacy guarantees, gives an upper bound on expected utility for this task. The non-private text classifier was developed by fine-tuning the pre-trained \texttt{distilRoBERTa} base model for a total of 30 epochs on our training set. The \texttt{AdamW} optimizer was used with the default learning rate=$5e^{-5}$ and batch size=8. 

\textbf{SWAG-PPM.}  The SWAG-PPM uses the same base pre-trained model, optimizer, and hyperparameters as the non-private model during its initial 10 epochs of fine-tuning.  However, it switches to using the \texttt{SGD} optimizer with a higher constant learning rate=0.01 for 20 epochs when exploring the parameter space required to estimate SWAG, resulting in 30 total epochs of training. The SWAG optimizer and hyperparameters noted above align with recommendations used in original works of \citet{maddox2019simple} and \citet{izmailov2018averaging}.  We take a total of 500 posterior draws from the 1st fitted SWAG to calculate the risk-based weights, using the hyperparameters of c=1 and g=0 for the linear mapping. We take 500 posterior draws from the 2nd fitted SWAG that incorporates the risk-based weights and pseudo-likelihood to calculate $\epsilon$ and a single draw to assess utility on the test set.

\textbf{SWAG-PPM Reweighted.} The reweighting extension of SWAG-PPM runs an additional round of training using a reweighting hyperparameter of k=0.95.  All other hyperparameters and design choices mimic those of SWAG-PPM.  The reweighted SWAG-PPM trains for a total of 30 epochs from the base \texttt{distilRoBERTa} model parameters.

\textbf{DP-SGD.} As a final private benchmark, we employ the DP-SGD mechanism using the same base pre-trained model and optimizer as the non-private model. As noted in the review by \citet{ponomareva2023dp}, the utility of DP-SGD and related gradient noise injection techniques are highly sensitive to their hyperparameter settings. As such, we use recommendations from \citet{li2021large} and \citet{yu2021differentially} for private fine-tuning of DP-SGD for transformer-based languages models like those used in our experiments.  Specifically, we use a batch size=512, learning rate=0.001, and gradient clipping norm=1. To compare the Renyi DP guarantee of DP-SGD to the ($\epsilon,\delta$)-DP of SWAG-PPM, we use the conversion provided in Proposition 3 of \citet{Mironov2017RnyiDP} and set a target $\epsilon$ for DP-SGD in Renyi DP to roughly match the $\epsilon$ for SWAG-PPM. Like the other benchmarks, DP-SGD was trained for a total of 30 epochs.

\subsection{Utility Metrics}

To evaluate model utility, we report the Macro F1 and Weighted F1 scores on the test set. These metrics are based on the F1 score, defined for class \( c \) as:
\[
F1_c = 2 \cdot \frac{\text{Precision}_c \cdot \text{Recall}_c}{\text{Precision}_c + \text{Recall}_c},
\]
where:
\[
\text{Precision}_c = \frac{TP_c}{TP_c + FP_c}, \quad \text{Recall}_c = \frac{TP_c}{TP_c + FN_c}.
\]
are constructed using counts of true positives (TP), false positives (FP), and false negatives (FN) within each class.

\subsubsection{Macro F1 Score}
The Macro F1 score is the unweighted mean of the F1 scores across all \( N \) classes:
\[
\text{Macro F1} = \frac{1}{N} \sum_{c=1}^N F1_c.
\]

\subsubsection{Weighted F1 Score}
The Weighted F1 score is the weighted average of the F1 scores, where weights \( w_c \) are proportional to the number of true instances \( n_c \) in each class:
\[
\text{Weighted F1} = \sum_{c=1}^N w_c \cdot F1_c, \quad \text{where } w_c = \frac{n_c}{\sum_{i=1}^N n_i}.
\]

These metrics offer complementary perspectives: Macro F1 treats all classes equally, while Weighted F1 accounts for class imbalance.

\section{Results}

Table \ref{table:1} summarizes the privacy and aggregate utility across benchmark models. For all models, the Weighted F1 is higher than the Macro F1, illustrating they perform relatively better on classes with more observations. SWAG-PPM exhibits only a modest drop in utility when compared to the non-private model, with slightly lower Weighted F1 (0.76 vs 0.77) and lower Macro F1 (0.44 vs 0.49).  This minimal loss in utility highlights a major benefit of the targeted nature of SWAG-PPM.  Since the mechanism downweights the contribution of a record proportional to its riskiness, it is able to mitigate the aggregate detrimental effects of noise injection while maintaining the same level of privacy.  In direct contrast, the DP-SGD model leads to a drastic drop in utility due to the global manner that noise is applied to the gradient which impacts the contribution of all records, irrespective of relative riskiness. As expected, the reweighted SWAG-PPM has improved utility when compared to SWAG-PPM though the magnitude of the improvement is modest when viewing the overall evaluation metrics due to the class imbalance (Weighted F1: 0.75; Macro F1: 0.45).

Though the utility metrics for DP-SGD are significantly worse than those of SWAG-PPM for the same approximate level of privacy, care should be taken given the subtle difference in privacy guarantees. Since the privacy comparison is challenging due to the absence of a finite sample estimate of $\delta$ for SWAG-PPM, we provide sensitivity analyses in Table \ref{tab:dp_sgd_swag} of Appendix \ref{app:delta} running DP-SGD under a wide range of $\delta$.  These additional investigations suggest that DP-SGD still greatly underperforms even under generous privacy relaxations (i.e., $\delta$=0.99).

\begin{table}[h!]
\centering
\caption{Privacy and Utility Comparison for Different Models}
\label{tab:privacy_utility}
\begin{tabular}{lcc|cc}
\toprule
\textbf{Model} & \multicolumn{2}{c|}{\textbf{Privacy}} & \multicolumn{2}{c}{\textbf{Utility}} \\
\cmidrule(lr){2-3} \cmidrule(lr){4-5}
& \textbf{Epsilon} & \textbf{Delta} & \textbf{F1 Weighted} & \textbf{F1 Macro} \\
\midrule
Non-Private & - & - & 0.76 & 0.49 \\
SWAG-PPM & 4.35 & $O(n^{-1/2})$ & 0.75 & 0.44 \\
SWAG-PPM (Reweighted) & 4.81 & $O(n^{-1/2})$ & 0.75 & 0.45 \\
DP-SGD & 4 & $10^{-4}$ & 0.08 & 0.03 \\
\bottomrule
\end{tabular}
\label{table:1}
\end{table}

\begin{figure}[h]
\centering
\includegraphics[width=\textwidth]{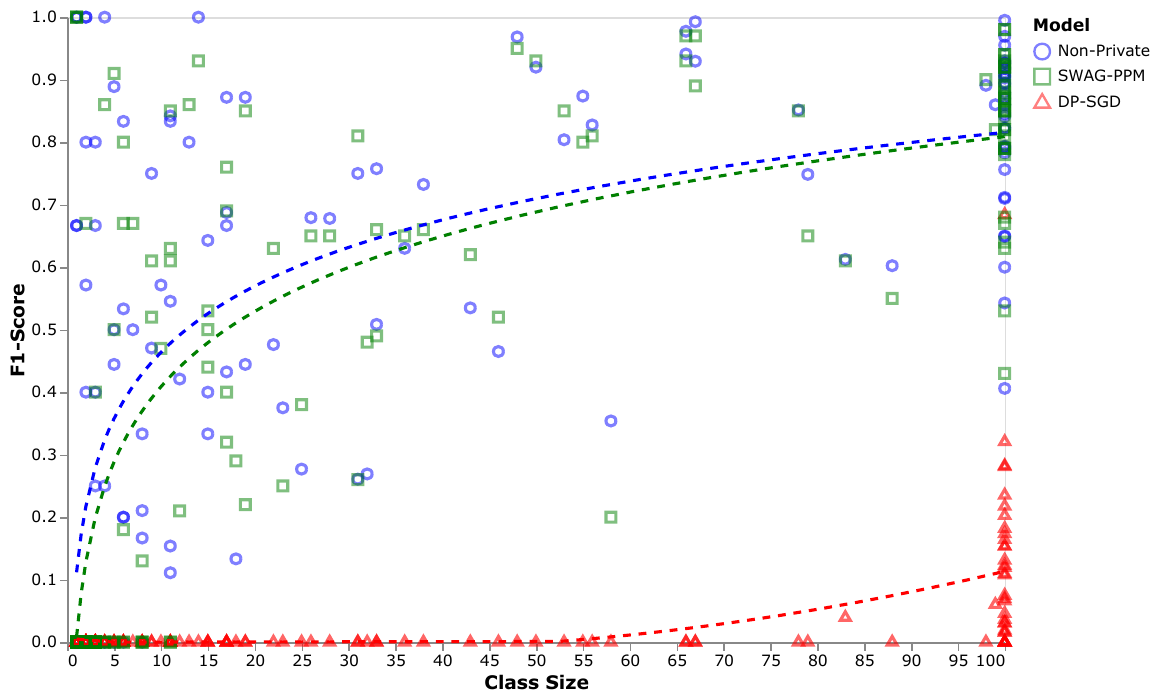}
\caption{SWAG-PPM vs DP-SGD Utility by Class Size}
\label{figure:2}
\end{figure}

To further investigate the utility comparison between the non-private model, SWAG-PPM, and DP-SGD, Figure \ref{figure:2} depicts the F1 scores by class size under each model. We see that the performance of DP-SGD is heavily influenced by the strong class imbalance, with non-zero F1 scores only appearing in categories with the most training observations. While SWAG-PPM also struggles with rare classes, it performs much better under these conditions, with examples of non-zero F1 scores across classes of different sizes.

The high utility of SWAG-PPM in this setting may be due to an interaction of the risk-based weights and the highly imbalanced class distribution. SWAG-PPM downweights observations that are poorly predicted by the model and exhibit high loss. Heavier downweighing is normally thought to reduce utility, since it forces the model to provide less attention to certain observations in the training set which maybe useful for future generalization.  However, if the high loss observations are highly concentrated in rare classes that are already challenging to predict in a non-private setting, then the negative impact of downweighting to utility is greatly tempered. 

\begin{figure}[h]
\centering
\includegraphics[width=\textwidth]{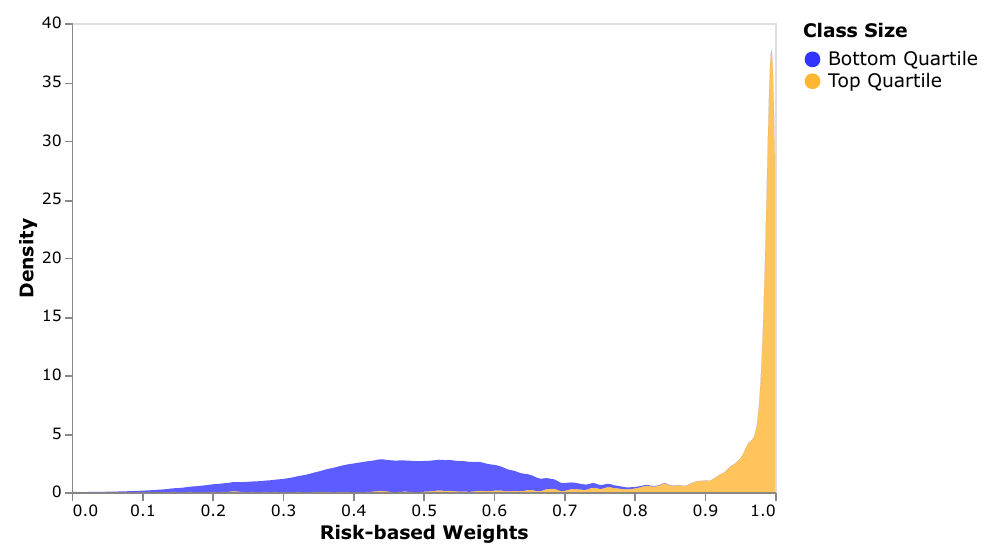}
\caption{Risk-based Weight Density Plot by Top and Bottom Class Size Quartiles}
\label{figure:3}
\end{figure}

To investigate this hypothesis, we aggregate the largest 25\% and smallest 25\% of classes together and compare the performance of the non-private and SWAG-PPM models on these aggregated classes (Table \ref{table:2}). On the aggregate class consisting of the top quartile of classes by size, both models exhibit the same test set weighted F1 (0.81) and macro F1 (0.81).  This is because most observations in these popular classes are only slightly downweighted (Figure \ref{figure:3}) and still contribute greatly to model training. In contrast, on the aggregate class consisting of the bottom quartile of classes by size, the non-private model outperforms SWAG-PPM on both weighted F1 (0.17 vs 0.06) and macro F1 (0.18 vs 0.08), showcasing the negative impact of the downweighting on utility. However, since the non-private model performance for rare classes is already poor, the drop in utility due to downweighting is minor when viewing the overall test set metrics.

\begin{table}[h!]
\centering
\caption{Model Performance by Class Size Quartile}
\label{tab:class_quartile}
\renewcommand{\arraystretch}{1.2}
\begin{tabular}{l l c c}
\toprule
\textbf{Model} & \textbf{Class Size Quartile} & \textbf{F1 Weighted} & \textbf{F1 Macro} \\ 
\midrule
Non-Private & Top   & 0.81 & 0.81  \\
SWAG-PPM & Top   & 0.81 & 0.81  \\
\midrule
Non-Private & Bottom & 0.17 & 0.18  \\
SWAG-PPM & Bottom & 0.06 & 0.08  \\
\bottomrule
\end{tabular}
\label{table:2}
\end{table}

\section{Discussion}
\label{sec:conc}

We have presented a scalable differential privacy mechanism for deep learning, SWAG-PPM, that exhibits both high utility and privacy protection under an approximate DP guarantee. In particular, we demonstrate that SWAG-PPM can perform well on datasets with high class imbalance and limited training data, including common tasks in official statistics such as workplace injury coding. We also show that the privacy-utility trade-off of SWAG can be tuned by use of reweighting, and provide supporting evidence that SWAG-PPM's high utility is driven by the selective downweighting combined with the low baseline accuracy of the models on rare classes.

Our findings corroborate prior literature showing that DP-SGD struggles with high class imbalance \citep{bagdasaryan2019differential, farrand2020neither} which is likely exacerbated under smaller training sizes \citep{rosenblatt2024differential}. This may be due to limitations of both the mechanism's components and privacy composition. \citet{farrand2020neither} hypothesized that DP-SGD struggles under class imbalance due to gradient clipping, which bounds the influence of outliers. Minority group examples, being fewer in each mini-batch, tend to have higher gradients that are more frequently clipped. This impairs the model's ability to learn effectively from the minority group, prioritizing accuracy improvements for the majority group instead. This effect is likely amplified by the mechanism's additive noise; since popular classes tend to have more observations per mini-batch, the noise impact may to some extent average out over the group, whereas noise added to the few clipped gradients of minority classes will disproportionately diminish their contribution.  Indeed, \citet{bagdasaryan2019differential} found that while implementing gradient clipping or additive noise independently only caused minor impacts to the underrepresented group, the combination of both in DP-SGD resulted in severe drops in accuracy. Though our results show that SWAG-PPM also tends to penalize observations from minority classes during training, the targeted downweighting of the per-record risk is a less severe suppression than uniform gradient clipping, and due to the differences in mechanism, SWAG-PPM does not suffer from the same amplified distortion from additive noise as DP-SGD.  

Another key difference between SWAG-PPM and DP-SGD is that mechanisms based on a model posterior distribution, like SWAG-PPM, have the useful property that the privacy guarantee does not depend on the number of posterior sampling iterations, provided that no intermediate sampling results are exposed. Privacy exposure arises from each release of a sampled parameter posterior value or an associated draw of synthetic data from the model's posterior predictive distribution conditioned on the parameter value. Unlike DP-SGD, this design offers a distinct advantage by focusing privacy accounting solely on the external releases, rather than requiring a composition over each training step.

Lastly, the stark difference in utility between SWAG-PPM and DP-SGD may also, in part, be due to the nature of the privacy guarantee of SWAG-PPM vs DP-SGD.  SWAG-PPM uses a notation of local sensitivity that provides protection to a specific private database as opposed to the global sensitivity of DP-SGD that provides protection independent of the risk of records within the private database. While using global sensitivity supports a broader privacy guarantee, mechanisms that use local sensitivity can provide more flexibility in privacy-utility trade-offs, since the focus is on the specific private database of interest instead of all possible private databases of a certain size. Furthermore, extending the PPM from local to global privacy guarantees has been explored by \cite{hu2025mechanisms} and can be applied to SWAG-PPM.    

This work provides various contributions to the literature. First, while not designed specifically to address private imbalanced learning, our results show that SWAG-PPM is far less impacted by class imbalance than standard approaches like DP-SGD. This suggests it may be a useful benchmark to compare to, or expand upon for, future methods in this nascent field of study. Second, our work carries forward the tradition of \citet{wang2015privacy} who first proposed the use of the posterior mechanism for DP Bayesian machine learning. To our knowledge, we are one of the first to extend the posterior mechanism to Bayesian deep learning and we further motivate its use by exploiting the privacy properties of the PPM. Finally, our application will hopefully inform the official statistics and survey research communities on the challenges of privatizing machine learning autocoders. More generally, our application provides a rare example of assessing differentially private fine-tuning of transformer models on severely imbalanced data. Prior work demonstrating the success of private fine-tuning of transformer-based languages models using DP-SGD \citep{li2021large, yu2021differentially} has primarily focused on balanced datasets, such as the Stanford Sentiment Treebank (SST-2) and Multi-Genre Natural Language Inference (MNLI) datasets. Our work calls into question the utility of this approach on imbalanced datasets for this important class of models.

\subsection{Limitations and Future Work}
\label{sec:limit-fw}

Our work has several limitations and natural avenues for future work.  First, although \citet{JMLR:v23:21-0936} provides an asymptotic bound of $\delta$ for the pseudo posterior mechanism (also in Appendix \ref{app:aDP}), they do not provide a means of quantifying $\delta$ for SWAG-PPM under any specific finite sample. Though this is a limitation of the approximate DP guarantee, we provide evidence in Appendix \ref{app:delta} that SWAG-PPM still performs better than DP-SGD under an extreme range of deltas (up to $\delta=0.99)$. Extending the PPM from local to global sensitivity for finite sample sizes has been investigated by \cite{hu2025mechanisms} who used a clipped log-likelihood. An extension to SWAG-PPM is the topic of ongoing research. Second, this analysis was run on a sample of the full OSHA dataset. Though it is  common in many research settings to have limited labeled training data available, our results provide less direct evidence into how the privacy and utility of SWAG-PPM and DP-SGD scale with increased training data. Future research could systematically explore the interplay of sample size, number of classes, class imbalance, and privacy guarantees for these methods.  Lastly, the SWAG approximation to the posterior distribution can generally be justified by the Bayesian central limit theorem when the sample size is much larger than the number of parameters. However, in a large and complex parameter space, as is the case for large language models, a single mode solution may be hard to justify. However, Gaussian distributions also provide the building blocks to non-parametric procedures such as mixture models, which allow for the approximation of much more complex underlying distributions. A potential avenue for future work is the use of SWAG to estimate a finite mixture of Gaussian distributions, such as MultiSWAG \citep{wilson2020bayesian}, which would then be used in combination with the PPM. 

\section*{Data Availability}
The data supporting this study are openly available at
\url{https://doi.org/10.6084/m9.figshare.28669604.v1}

\section*{Funding}
The authors gratefully acknowledge support from the ASA/NSF/BLS Senior Research Fellow program.

\section*{Conflict of Interest}
The authors have no competing interests to declare.

\bibliography{A_reference}

\appendix
\section{Detailed Utility Metrics}

\begin{singlespace}
\begin{longtable}{|c|c|c|c|c|}
\caption{Test set Class Sizes and Class-specific F1 scores}
\\ \hline
\multirow{2}{*}{\textbf{OIICS Code}} & \multirow{2}{*}{\textbf{Class Size (Test)}} & \multicolumn{3}{|c|}{\textbf{F1-Score}} \\ \cline{3-5}
 & & \textbf{Non-Private} & \textbf{SWAG-PPM} & \textbf{DP-SGD} \\ \hline
1720 & 100 & 0.91 & 0.87 & 0.02 \\ \hline
1725 & 100 & 0.92 & 0.94 & 0.32 \\ \hline
1728 & 100 & 0.71 & 0.78 & 0.17 \\ \hline
1729 & 100 & 0.76 & 0.79 & 0.07 \\ \hline
124 & 100 & 1.00 & 0.98 & 0.18 \\ \hline
1830 & 100 & 0.65 & 0.68 & 0.07 \\ \hline
1831 & 100 & 0.91 & 0.87 & 0.11 \\ \hline
1834 & 100 & 0.93 & 0.92 & 0.05 \\ \hline
1839 & 100 & 0.41 & 0.43 & 0.00 \\ \hline
1960 & 100 & 0.54 & 0.53 & 0.02 \\ \hline
1963 & 100 & 0.82 & 0.82 & 0.28 \\ \hline
1971 & 100 & 0.90 & 0.88 & 0.11 \\ \hline
1972 & 100 & 0.60 & 0.63 & 0.00 \\ \hline
132 & 100 & 0.86 & 0.85 & 0.15 \\ \hline
133 & 100 & 0.71 & 0.67 & 0.07 \\ \hline
134 & 100 & 0.97 & 0.98 & 0.68 \\ \hline
10 & 100 & 0.79 & 0.81 & 0.22 \\ \hline
143 & 100 & 0.84 & 0.86 & 0.00 \\ \hline
160 & 100 & 0.79 & 0.82 & 0.12 \\ \hline
161 & 100 & 0.79 & 0.79 & 0.00 \\ \hline
162 & 100 & 0.92 & 0.91 & 0.00 \\ \hline
193 & 100 & 0.89 & 0.92 & 0.24 \\ \hline
194 & 100 & 0.65 & 0.64 & 0.00 \\ \hline
1212 & 100 & 0.88 & 0.89 & 0.03 \\ \hline
1231 & 100 & 0.82 & 0.84 & 0.03 \\ \hline
111 & 100 & 0.78 & 0.79 & 0.02 \\ \hline
1311 & 100 & 0.93 & 0.93 & 0.13 \\ \hline
1312 & 100 & 0.87 & 0.87 & 0.20 \\ \hline
1510 & 100 & 0.89 & 0.85 & 0.28 \\ \hline
1520 & 100 & 0.85 & 0.85 & 0.16 \\ \hline
1522 & 100 & 0.91 & 0.89 & 0.15 \\ \hline
1523 & 100 & 0.96 & 0.94 & 0.12 \\ \hline
1530 & 100 & 0.93 & 0.92 & 0.04 \\ \hline
181 & 99 & 0.86 & 0.82 & 0.06 \\ \hline
1233 & 98 & 0.89 & 0.90 & 0.00 \\ \hline
130 & 88 & 0.60 & 0.55 & 0.00 \\ \hline
1961 & 83 & 0.61 & 0.61 & 0.04 \\ \hline
1973 & 79 & 0.75 & 0.65 & 0.00 \\ \hline
1681 & 78 & 0.85 & 0.85 & 0.00 \\ \hline
1211 & 67 & 0.93 & 0.89 & 0.00 \\ \hline
1532 & 67 & 0.99 & 0.97 & 0.00 \\ \hline
2331 & 66 & 0.94 & 0.93 & 0.00 \\ \hline
1533 & 66 & 0.98 & 0.97 & 0.00 \\ \hline
189 & 58 & 0.35 & 0.20 & 0.00 \\ \hline
1966 & 56 & 0.83 & 0.81 & 0.00 \\ \hline
1232 & 55 & 0.87 & 0.80 & 0.00 \\ \hline
141 & 53 & 0.80 & 0.85 & 0.00 \\ \hline
1833 & 50 & 0.92 & 0.93 & 0.00 \\ \hline
1721 & 48 & 0.97 & 0.95 & 0.00 \\ \hline
1999 & 46 & 0.47 & 0.52 & 0.00 \\ \hline
120 & 43 & 0.53 & 0.62 & 0.00 \\ \hline
1722 & 38 & 0.73 & 0.66 & 0.00 \\ \hline
185 & 36 & 0.63 & 0.65 & 0.00 \\ \hline
1968 & 33 & 0.51 & 0.49 & 0.00 \\ \hline
1689 & 33 & 0.76 & 0.66 & 0.00 \\ \hline
1230 & 32 & 0.27 & 0.48 & 0.00 \\ \hline
180 & 31 & 0.26 & 0.26 & 0.00 \\ \hline
1512 & 31 & 0.75 & 0.81 & 0.00 \\ \hline
138 & 28 & 0.68 & 0.65 & 0.00 \\ \hline
1590 & 26 & 0.68 & 0.65 & 0.00 \\ \hline
1832 & 25 & 0.28 & 0.38 & 0.00 \\ \hline
150 & 23 & 0.38 & 0.25 & 0.00 \\ \hline
1849 & 22 & 0.48 & 0.63 & 0.00 \\ \hline
1978 & 19 & 0.44 & 0.22 & 0.00 \\ \hline
1513 & 19 & 0.87 & 0.85 & 0.00 \\ \hline
1979 & 18 & 0.13 & 0.29 & 0.00 \\ \hline
1962 & 17 & 0.67 & 0.69 & 0.00 \\ \hline
1965 & 17 & 0.69 & 0.40 & 0.00 \\ \hline
5111 & 17 & 0.43 & 0.32 & 0.00 \\ \hline
1319 & 17 & 0.87 & 0.76 & 0.00 \\ \hline
129 & 15 & 0.33 & 0.53 & 0.00 \\ \hline
190 & 15 & 0.40 & 0.44 & 0.00 \\ \hline
1521 & 15 & 0.64 & 0.50 & 0.00 \\ \hline
1711 & 14 & 1.00 & 0.93 & 0.00 \\ \hline
1131 & 13 & 0.80 & 0.86 & 0.00 \\ \hline
169 & 12 & 0.42 & 0.21 & 0.00 \\ \hline
191 & 11 & 0.15 & 0.00 & 0.00 \\ \hline
1120 & 11 & 0.55 & 0.61 & 0.00 \\ \hline
1121 & 11 & 0.83 & 0.85 & 0.00 \\ \hline
110 & 11 & 0.11 & 0.00 & 0.00 \\ \hline
1531 & 11 & 0.84 & 0.63 & 0.00 \\ \hline
128 & 10 & 0.57 & 0.47 & 0.00 \\ \hline
1974 & 9 & 0.75 & 0.52 & 0.00 \\ \hline
148 & 9 & 0.47 & 0.61 & 0.00 \\ \hline
1964 & 8 & 0.17 & 0.00 & 0.00 \\ \hline
1969 & 8 & 0.00 & 0.00 & 0.00 \\ \hline
1210 & 8 & 0.33 & 0.00 & 0.00 \\ \hline
1238 & 8 & 0.21 & 0.13 & 0.00 \\ \hline
1130 & 7 & 0.50 & 0.67 & 0.00 \\ \hline
1967 & 6 & 0.20 & 0.00 & 0.00 \\ \hline
2330 & 6 & 0.53 & 0.67 & 0.00 \\ \hline
2361 & 6 & 0.83 & 0.80 & 0.00 \\ \hline
2811 & 6 & 0.20 & 0.18 & 0.00 \\ \hline
9999 & 6 & 0.00 & 0.00 & 0.00 \\ \hline
7 & 5 & 0.00 & 0.00 & 0.00 \\ \hline
1822 & 5 & 0.44 & 0.00 & 0.00 \\ \hline
5112 & 5 & 0.50 & 0.50 & 0.00 \\ \hline
1511 & 5 & 0.89 & 0.91 & 0.00 \\ \hline
1829 & 4 & 0.00 & 0.00 & 0.00 \\ \hline
142 & 4 & 1.00 & 0.86 & 0.00 \\ \hline
230 & 4 & 0.25 & 0.00 & 0.00 \\ \hline
1591 & 4 & 0.00 & 0.00 & 0.00 \\ \hline
8 & 3 & 0.00 & 0.00 & 0.00 \\ \hline
1950 & 3 & 0.40 & 0.00 & 0.00 \\ \hline
1992 & 3 & 0.80 & 0.00 & 0.00 \\ \hline
2431 & 3 & 0.00 & 0.00 & 0.00 \\ \hline
2433 & 3 & 0.67 & 0.40 & 0.00 \\ \hline
6210 & 3 & 0.00 & 0.00 & 0.00 \\ \hline
518 & 3 & 0.25 & 0.00 & 0.00 \\ \hline
1129 & 3 & 0.00 & 0.00 & 0.00 \\ \hline
1219 & 3 & 0.00 & 0.00 & 0.00 \\ \hline
1310 & 3 & 0.00 & 0.00 & 0.00 \\ \hline
118 & 3 & 0.00 & 0.00 & 0.00 \\ \hline
1592 & 3 & 0.00 & 0.00 & 0.00 \\ \hline
1680 & 3 & 0.00 & 0.00 & 0.00 \\ \hline
1821 & 2 & 0.00 & 0.00 & 0.00 \\ \hline
1840 & 2 & 0.00 & 0.00 & 0.00 \\ \hline
1841 & 2 & 0.80 & 0.00 & 0.00 \\ \hline
125 & 2 & 0.00 & 0.00 & 0.00 \\ \hline
1952 & 2 & 0.40 & 0.00 & 0.00 \\ \hline
2332 & 2 & 1.00 & 0.00 & 0.00 \\ \hline
2810 & 2 & 0.00 & 0.00 & 0.00 \\ \hline
5110 & 2 & 0.00 & 0.00 & 0.00 \\ \hline
5118 & 2 & 0.00 & 0.00 & 0.00 \\ \hline
5159 & 2 & 0.57 & 0.67 & 0.00 \\ \hline
5169 & 2 & 0.00 & 0.00 & 0.00 \\ \hline
6119 & 2 & 0.00 & 0.00 & 0.00 \\ \hline
139 & 2 & 0.00 & 0.00 & 0.00 \\ \hline
1221 & 2 & 0.00 & 0.00 & 0.00 \\ \hline
1593 & 2 & 1.00 & 0.00 & 0.00 \\ \hline
1733 & 1 & 0.00 & 1.00 & 0.00 \\ \hline
1820 & 1 & 0.00 & 0.00 & 0.00 \\ \hline
1951 & 1 & 0.00 & 0.00 & 0.00 \\ \hline
1991 & 1 & 0.00 & 0.00 & 0.00 \\ \hline
2362 & 1 & 0.00 & 0.00 & 0.00 \\ \hline
2491 & 1 & 1.00 & 0.00 & 0.00 \\ \hline
2492 & 1 & 1.00 & 1.00 & 0.00 \\ \hline
2813 & 1 & 0.67 & 1.00 & 0.00 \\ \hline
2919 & 1 & 0.00 & 0.00 & 0.00 \\ \hline
5119 & 1 & 0.00 & 0.00 & 0.00 \\ \hline
5150 & 1 & 0.00 & 0.00 & 0.00 \\ \hline
5158 & 1 & 0.00 & 0.00 & 0.00 \\ \hline
140 & 1 & 0.00 & 0.00 & 0.00 \\ \hline
178 & 1 & 0.00 & 0.00 & 0.00 \\ \hline
30 & 1 & 0.00 & 0.00 & 0.00 \\ \hline
192 & 1 & 0.00 & 0.00 & 0.00 \\ \hline
310 & 1 & 0.00 & 0.00 & 0.00 \\ \hline
1139 & 1 & 0.00 & 0.00 & 0.00 \\ \hline
1220 & 1 & 0.00 & 0.00 & 0.00 \\ \hline
1229 & 1 & 0.00 & 0.00 & 0.00 \\ \hline
119 & 1 & 0.00 & 0.00 & 0.00 \\ \hline
1599 & 1 & 0.00 & 0.00 & 0.00 \\ \hline
1712 & 1 & 0.67 & 0.00 & 0.00 \\ \hline
\end{longtable}
\label{tab:detailed-utility}
\end{singlespace}

\section{Impact of $\delta$ on DP-SGD Utility}\label{app:delta}
Since our $\delta$ estimate for SWAG-PPM is approximate (asymptotic), for a fair comparison we run DP-SGD under more optimistic settings for $\delta$ including extremely large values. Table \ref{tab:dp_sgd_swag} demonstrates that utility does increase for DP-SGD for a given target epsilon as $\delta$ is relaxed. However, even under an extreme setting of $\delta = 0.99$, performance is still much worse for DP-SGD compared to the SWAG-PPM

\begin{table}[h!]
\centering
\caption{Comparison of SWAG-PPM and DP-SGD Across a Range of $\delta$}
\label{tab:dp_sgd_swag}
\renewcommand{\arraystretch}{1.2}
\begin{tabular}{l c c c c}
\toprule
\textbf{Method} & \textbf{Target Epsilon} & \textbf{Delta} & \textbf{F1 Weighted} & \textbf{F1 Macro} \\
\midrule
DP-SGD & 4 & 0.001 & 0.08 & 0.03 \\
DP-SGD & 4 & 0.01  & 0.08 & 0.03 \\
DP-SGD & 4 & 0.1   & 0.16 & 0.06 \\
DP-SGD & 4 & 0.99  & 0.35 & 0.13 \\
SWAG-PPM   & 4.35 & \(O(n^{-1/2})\) & 0.75 & 0.44 \\
\bottomrule
\end{tabular}
\end{table}

\section{Asymptotic Bounds}\label{app:aDP}
We recall that the local sensitivity is defined as the maximum risk or log-likelihood contribution observed over the given data set $D$ and realized draws $\theta_S$ from the (pseudo) posterior:
\[
\Delta_{\bm{\alpha},\mathbf{D}} = \max_{\theta_s \sim \xi^{\alpha}(\theta|D)} \left\{\inf \left\{w: \abs{\ell^{\bm{\alpha}(D_i)}_{\theta}(D_i)} \leq w, i \in \{1, \ldots n\}, D_i \sim P(D|\theta_0) \right\} \right\}
\]
In contrast, the global sensitivity is defined as over the entire support of the (pseudo) posterior and the entire class of possible data sets for which $D$ belongs:
\[
\Delta_{\bm{\alpha}} = \sup_{\theta \sim \xi^{\alpha}(\theta|D)} \left\{\inf \left\{w: \abs{\ell^{\bm{\alpha}(D_i)}_{\theta}(D_i)} \leq w, \forall D_i \in \mathbf{D},~\forall \mathbf{D} \in \mathcal{D}^{n} \right\} \right\}
\]
It is clear that the local sensitivity $\Delta_{\bm{\alpha},\mathbf{D}}$ is an underestimate of the global sensitivity $\Delta_{\bm{\alpha}}$ because
\begin{enumerate}
    \item A maximum over the posterior draws will always grow if more draws are taken. For example if we increase the SWAG draws from 500 to 50,000, our estimate $\Delta_{\bm{\alpha},\mathbf{D}}$ cannot decrease.
    \item In the case when $D$ may contain variables that are continuous and unbounded, there may always be an new unobserved record $D_i$ that is riskier than any observed ones. For example, if we receive new survey responses with unique text that has not been seen in training before, this may lead to a poorly predicted value with a log-likelihood closer to 0 than any in our training set.
\end{enumerate}
The first point can be addressed by an asymptotic argument.
In section 4.3 of \citet{JMLR:v23:21-0936} they justify that asymptotically 
$P(|\Delta_{\bm{\alpha}} - \Delta_{\bm{\alpha},\mathbf{D}}| > 0) \rightarrow 0$ as the sample size grows and the posterior distribution for parameters $\theta$ concentrates around a single point $\theta^*$ that minimizes the Kullback-Liebler (KL) divergence between the true model (e.g. all $\alpha = 1$) and the best achievable model under mis-specification ($\alpha_i < 1)$. At a high level the argument in \citet{JMLR:v23:21-0936} is the following
\begin{itemize}
    \item Theorem: Contraction of the $\alpha-$ weighted pseudo-posterior to a point $\theta^*$.
    \begin {itemize}
        \item Assumption: Sufficient prior mass covers the true $\theta_0$.
        \item Assumption: The number of down-weighted records (those with $\alpha_i < 1$) grows at a slower rate than the overall sample size (e.g. a rate of $\sqrt{n}$).
    \end{itemize}
    \item Let $\Delta_{\bm{\alpha},\mathbf{D}}^{\infty}$ be the limiting sensitivity evaluated at the single parameter set $\theta^*$:
    \[
\Delta_{\bm{\alpha},\mathbf{D}}^{\infty} =  \inf \left\{w: \abs{\ell^{\bm{\alpha}(D_i)}_{\theta^*}(D_i)} \leq w, i \in \{1, \ldots n\}, D_i \sim P(D|\theta_0) \right\} 
    \]
    \begin{itemize}
        \item The global sensitivity and limiting sensitivity converge with respect to the true data generating distribution: $P_{\theta_0}(|\Delta_{\bm{\alpha}} - \Delta_{\bm{\alpha},\mathbf{D}}^{\infty}| > 0) \rightarrow 0$.
        \item The local sensitivity and the limiting sensitivity converge converge with respect to the true data generating distribution:
        $P_{\theta_0}(|\Delta_{\bm{\alpha},\mathbf{D}} - \Delta_{\bm{\alpha},\mathbf{D}}^{\infty}| > 0) \rightarrow 0$.
        \item   Then the local and global sensitivity converge converge with respect to the true data generating distribution:
        $P_{\theta_0}(|\Delta_{\bm{\alpha}} - \Delta_{\bm{\alpha},\mathbf{D}}| > 0) \rightarrow 0$.
    \end{itemize}
\end{itemize}

The second point is mitigated by the use of record-indexed weights, where riskier records are down-weighted. If a new riskier value of $D_i$ was used instead of coming from the observed set, it would receive a greater down-weighting and not lead to a net increase in $\Delta_{\bm{\alpha},\mathbf{D}}$. The re-weighting work \citep{savitsky2020re} might address this more efficiently (e.g. faster convergence) as it adaptively adjusts the down-weighting. 
\end{document}